\documentclass[final]{cvpr}

\usepackage{times}
\usepackage{epsfig}
\usepackage{graphicx}
\usepackage{amsmath}
\usepackage{amssymb}

\usepackage{float}
\usepackage{graphics} 
\usepackage{epsfig} 
\usepackage{mathptmx} 
\usepackage{import}
\usepackage{colortbl}
\usepackage{wrapfig,lipsum}
\usepackage{graphicx}
\usepackage{booktabs}
\usepackage{multirow}
\usepackage{placeins}
\usepackage{gensymb}
\usepackage{xcolor}
\usepackage{soul}
\usepackage{textcomp}
\usepackage[ruled,linesnumbered, noend]{algorithm2e}
\usepackage{yfonts}
\usepackage{atbegshi,picture}
\usepackage{cite}
\usepackage[font=small,labelfont=bf,tableposition=top]{caption}
\DeclareMathOperator*{\argminB}{argmin}   

\newenvironment{myitem}{\begin{list}{$\bullet$}
{\setlength{\itemsep}{-0pt}
\setlength{\topsep}{0pt}
\setlength{\labelwidth}{0pt}
\setlength{\leftmargin}{10pt}
\setlength{\parsep}{-0pt}
\setlength{\itemsep}{0pt}
\setlength{\partopsep}{0pt}}}%
{\end{list}}

\definecolor{Gray}{gray}{0.85}
\newcolumntype{a}{>{\columncolor{Gray}}c}
\usepackage{hyperref}

\def\cvprPaperID{14} 
\def\confYear{CVPR 2021}

\begin{document}


\title{Data-driven 6D Pose Tracking\\
by Calibrating Image Residuals in Synthetic Domains}

\author{Bowen Wen\\
Rutgers University\\
{\tt\small bw344@scarletmail.rutgers.edu}
\and
Chaitanya Mitash\\
Amazon\\
{\tt\small cmitash@amazon.com}
\and
Kostas Bekris \\
Rutgers University\\
{\tt\small kostas.bekris@cs.rutgers.edu}
}

\AtBeginShipoutNext{\AtBeginShipoutUpperLeft{%
  \put(\dimexpr\paperwidth-0.5cm\relax,-0.7cm){\makebox[0pt][r]{\framebox{CVPR 2021 Workshop on 3D Vision and Robotics}}}%
}}

\maketitle

\begin{abstract}
Tracking the 6D pose of objects in video sequences is important for  robot manipulation. This work presents $se(3)$-TrackNet, a data-driven optimization approach for long-term, 6D pose tracking. It aims to identify the optimal relative pose given the current RGB-D observation and a synthetic image conditioned on the previous best estimate and the object's model. The key contribution in this context is a novel neural network architecture, which appropriately disentangles the feature encoding to help reduce domain shift, and an effective 3D orientation representation via Lie Algebra. Consequently, even when the network is trained solely with synthetic data can work effectively over real images. Comprehensive experiments over multiple benchmarks show $se(3)$-TrackNet achieves consistently robust estimates and outperforms alternatives, even though they have been trained with real images. The approach runs in real time at 90.9Hz. Code,  data  and  supplementary  video  for  this  project  are  available  at\footnote{This work has been previously accepted to appear at IEEE IROS 2020 \cite{wense3tracknet}. This is a CVPRW'21 paper on 3D Vision and Robotics.} \small{\href{https://github.com/wenbowen123/iros20-6d-pose-tracking}{https://github.com/wenbowen123/iros20-6d-pose-tracking}}
\end{abstract}

\section{Introduction}
Robotic tasks, such as object manipulation, often require to track the pose of an object.  Pose estimation from a single snapshot can initiate a manipulation pipeline \cite{xiang2017posecnn,wang2019densefusion}. Purposeful manipulation, however, such as placement \cite{mitash2020task,andrewbwrss} frequently involves dynamic object pose change.  Some pose estimation approaches are relatively fast and can re-estimate pose from scratch for every frame \cite{tremblay2018deep, wang2019densefusion, sundermeyer2018implicit,wen2020robust}. This can be redundant, however, and often leads to less coherent estimations over consecutive frames, which negatively impact manipulation.

Temporal tracking of object poses over sequences of images can greatly improve speed while maintain or even improve pose
quality \cite{wense3tracknet,Wthrich2013ProbabilisticOT, schmidt2014dart,stoiber2020sparse,magnusobject}. Nevertheless, many traditional methods that depend on hand-crafted likelihood functions and features, require extensive hyper-parameter tuning when adapted to novel object categories or environments. On the other hand, data-driven techniques \cite{li2018deepim,garon2017deep} require 
real-world training data, which are difficult to acquire and
label in the context of 6D poses.

\begin{figure}[t]
\vspace{-0.15in}
  \centering
  \includegraphics[width=0.45\textwidth,height=0.2\textwidth]{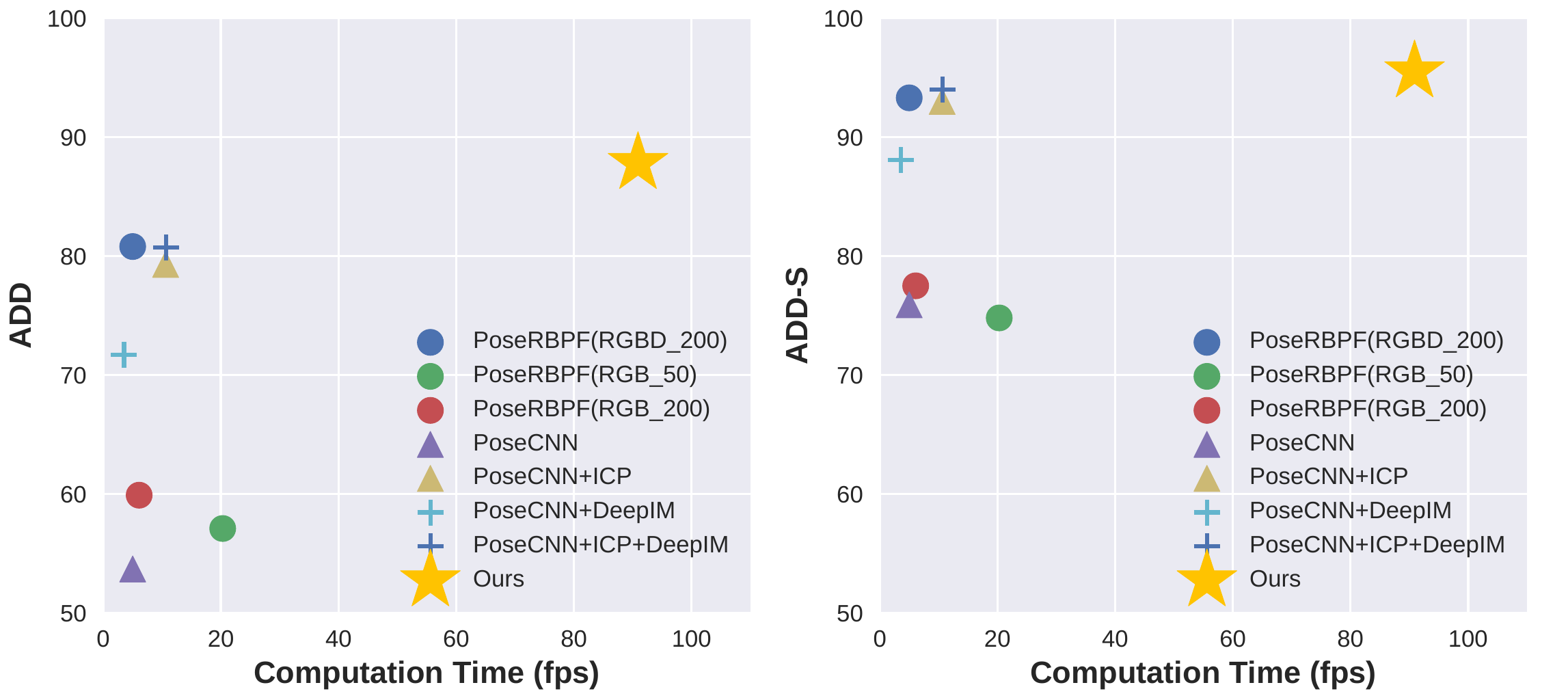}
  \includegraphics[width=0.45\textwidth]{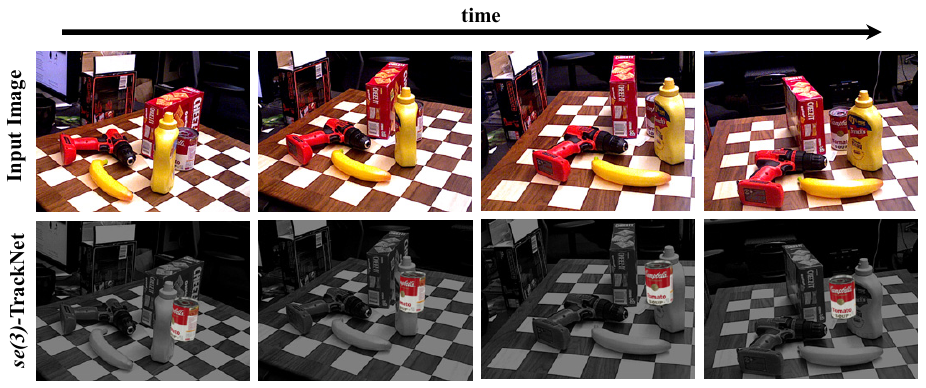}
  \vspace{-0.15in}
\caption{\scriptsize \textbf{Top:} Performance w.r.t. computation time evaluated on the YCB-Video dataset according to the area under the curve (AUC) metric for the ADD and ADD-S objectives \cite{xiang2017posecnn}. The proposed approach achieves more accurate tracking and is significantly faster than alternatives. \textbf{Bottom:} Pose predicted by the $se(3)$-TrackNet without any re-initialization, which is able to recover from complete occlusion.}
\label{fig:acc_speed}
\vspace{-0.3in}
\end{figure}

This work proposes a data-driven optimization strategy to keep long-term track of an object's 6D pose robustly. The
contributions are the following:

\noindent 1. A novel deep neural network for 6D object pose tracking, where a smart feature-encoding disentanglement technique enables more efficient sim-to-real transfer.

\noindent 2. A Lie Algebra representation of 3D orientations, which allows effective learning of the residual pose transforms given a proper loss function. 

\noindent 3. A training pipeline over synthetic data that employs 
domain randomization \cite{tobin2017domain} in the context of pose tracking. 

\noindent 4. A novel benchmarking dataset for 6D pose tracking in the context of multiple different robotics manipulation tasks, with groundtruth 6D pose annotation.

Experiments indicate that the proposed network achieves state-of-art
results on the YCB-Video benchmark without re-initialization in contrast to prior work \cite{li2018deepim,deng2019poserbpf}. It is also significantly faster at 90.9Hz, as shown in Fig. \ref{fig:acc_speed}.

\section{Related Work}
\noindent \textbf{6D Object Pose Estimation:} Learning-based techniques have shown promise in directly regressing the 6D object pose from image data \cite{mitash2019scene,
issac2016depth, sundermeyer2018implicit,li2019cdpn,he2020pvn3d}. Nevertheless, given the complexity of the 6D
challenge, a large amount of pose annotated training data is required
to achieve satisfactory results. This is often more
challenging than labeling for object classification or detection. Given that a pose is re-estimated in every frame,
estimation techniques often trade-off speed for accuracy
\cite{xiang2017posecnn} while still insufficient for real-time application. This is desirable for manipulation that often involves dynamic scene updates. In contrast,  $se(3)$-TrackNet exploits
temporal information to achieve higher accuracy and faster response than 
state-of-art single-image pose estimation methods while using only synthetic data for training.

\noindent \textbf{6D Object Pose Tracking:} For setups where CAD object models are available, prior approaches \cite{choi2013rgb, Wthrich2013ProbabilisticOT, issac2016depth} harness GPUs for probabilistic filtering. They often require, however, hand-crafted feature design and carefully tuned hyper-parameters, which do not generalize easily. Recent work \cite{deng2019poserbpf}
proposed a data-driven Rao-Blackwellized particle filter, achieving promising results on the YCB Video benchmark. It suffers, however, from severe occlusions and requires re-initialization from costly pose estimation. Another effort computes relative transformations between consecutive frames by optimizing for the discrepancy between consecutive observations \cite{schmidt2014dart, pauwels2015simtrack,
  joseph2015versatile, zhong2019robust, tjaden2018region}. The most related effort to the current paper leverages the FlowNetSimple network \cite{fischer2015flownet} for learning 6D pose tracking. It requires, however, occasional re-initialization and has to be trained at least partially with real data.

\section{Approach of $se(3)$-TrackNet}

\begin{figure}[h]
  \vspace{-0.1in}
  \centering
  \includegraphics[width=0.45\textwidth]{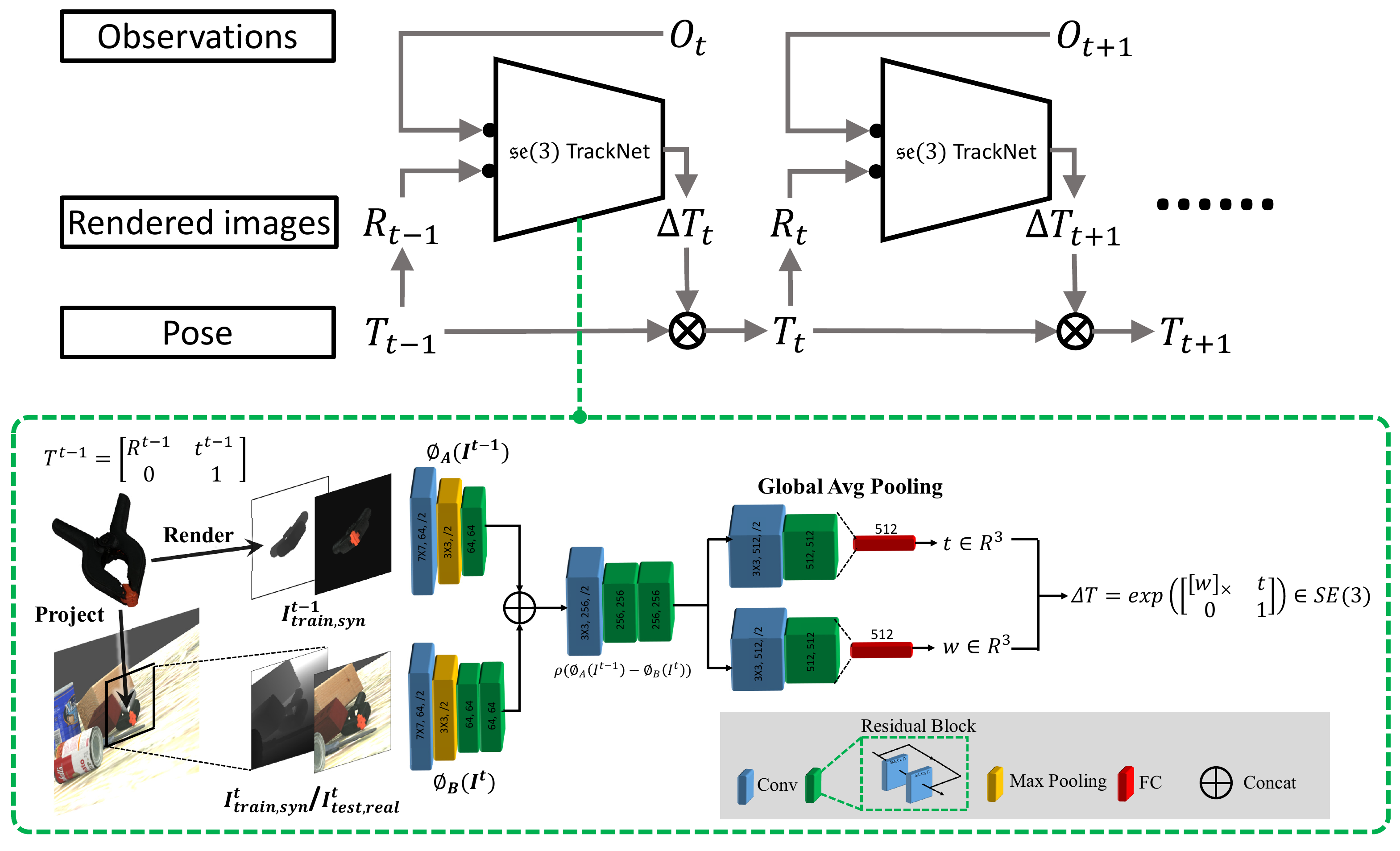}
  \vspace{-0.1in}
  \caption{\scriptsize Overview: At any given time $t$ the current observation $O_t$ and the rendering $R_{t-1}$ of the object model based on the previously computed pose $T_{t-1}$ are passed to the $se(3)$-TrackNet via two separate feature encoders $\phi_B$ and $\phi_A$ respectively. Both inputs are synthetic during training while at test time, the input to $\phi_B$ is a real observed image. The encoders' outputs are concatenated and used to predict the relative pose $\Delta T_{t}$ between the two images, with decoupled translation and rotational. $\Delta T_{t}$ is then combined with $T_{t-1}$ to compute $T_t$. The same process repeats along the video.}
  \label{fig:overview}
  \vspace{-0.1in}
\end{figure}

The objective is to compute the 6D pose of an object $T_t \in SE(3)$ at any time $t > 0$, given as input:
\begin{myitem}
    \item a 3D {\tt CAD} model of the object,
    \item its initial pose, $T_0 \in SE(3) $, computed by any single-image based 6D pose estimation technique, and a
    \item sequence of RGB-D images $O_{\tau }, \tau \in \{0,1,...,t-1\}$ from previous time stamps and the current observation $O_t$.
\end{myitem}
This work proposes a data-driven optimization technique to track the 6D object pose. The cost function for the optimization is encoded and learned by a novel neural network architecture, trained with only synthetically generated data. In every time step, it computes a relative pose $\Delta T_{\tau}$ from the previous frame as indicated in Fig. \ref{fig:overview}. 

\subsection{Tracking on $SE(3)$ Manifolds with Residuals}
\label{sec:se3}

Optimization in this domain operates over cost functions defined for the object poses $\bar{\xi}, \xi$ that measure the discrepancy $\epsilon$ between the features extracted from the images:
$$\varepsilon = \rho(\phi_{I_1} (\bar{\xi}) - \phi _{I_2}(\xi)),$$ where $\rho$ is a predefined robust loss function, and $\phi(\cdot)$ can be direct pixel intensity values, such as in \cite{engel2017direct}, point-to-point discrepancy or its variations \cite{besl1992method}, pre-designed features \cite{hebert2012combined} or the combinations from any of the above \cite{pauwels2015simtrack}. 

Given the current observation $O_t$, and the pose computed in the previous timestamp $\xi_{t-1}$, the goal of this work is to find a relative transformation $\Delta \xi$ that takes the object from $\xi_{t-1}$ to the pose captured by the current observation. This can be formulated as an optimization problem. Let $R$ denote the image corresponding to the object model's rendering at the given pose. Then, the optimal relative transform is:
\vspace{-0.1in}
$$
\Delta \xi^* = \argminB_{\Delta \xi} \{\rho(\phi_{O_{t}}(\xi_t) - \phi_R(\xi_{t-1} \boxplus \Delta \xi))\}
\vspace{-0.1in}
$$
An analytical solution is to perform Gauss-Newton optimization, which often requires appropriate choice of a robust cost function and hand-crafted features. Another problem arises when different modalities are involved and suitable tuning of relative weights struggles to generalize to new scenarios. Instead, this work proposes a novel neural network architecture that implicitly learns to calibrate the residual between the features extracted from the current observation and the rendered image conditioned on previous pose estimate to resolve the relative transform in the tangent space $\Delta \xi \in se(3)$. 

\subsection{Neural Network Design}
The proposed neural network is shown in Fig. \ref{fig:overview}. The network takes as input a pair of images, $I^{t-1}$: rendered from the previous pose estimation, and $I^{t}$: the current observation. The images are 4-channel RGB-D data. During training, both inputs are synthetically generated images $\phi(I_{syn,train}^{t-1};I_{syn,train}^{t})$, while for testing the current timestamp input comes from a real sensor, $\phi(I_{syn,test}^{t-1};I_{real,test}^{t})$. The $se(3)$-TrackNet uses two separate input branches for $I^{t-1}$ and $I^{t}$. The weights of the feature encoders are not shared so as to disentangle feature encoding. This is different from related work \cite{li2018deepim}, where the two images are concatenated into a single input. A shared feature extractor worked in the previous work when both real and synthetic data are available during training. The property of the latent space $\phi(I_{syn,train}^{t-1};I_{real,train}^{t})$ could still be partly preserved when tested on real world test scenarios $\phi(I_{syn,test}^{t-1};I_{real,test}^{t})$. This representation, however, does not generalize to training exclusively on synthetic data. 

In particular, let's denote the latent space features trained on purely synthetic data as $\phi_A(I_{syn,train}^{t-1})$ and $\phi_B(I_{syn,train}^{t})$. When tested on real world data, the latent space features are $\phi_A(I_{syn,test}^{t-1})$ and $\phi_B(I_{real,test}^{t})$. By this feature encoding disentanglement, domain gap reduces to be between $\phi_B(I_{syn,train}^{t})$ and $\phi_B(I_{real,test}^{t})$, while $\phi_A(I_{syn,train}^{t-1})$ and $\phi_A(I_{syn,test}^{t-1})$ can be effortlessly aligned between the training and test phase without the need for tackling the domain gap problem. 

A relative transform can be predicted by the network via end-to-end training. The transformation is represented by Lie algebra as $\Delta \xi = (t,w)^T  \in se(3)$, where the prediction of $w$ and $t$ are disentangled into separate branches and trained by $L_2$ loss: 
\vspace{-0.15in}
$$
L=\lambda_1||w-\bar{w}||_2 + \lambda_2||t-\bar{t}||_2
\vspace{-0.1in}
$$
\noindent where $\lambda_1$ and $\lambda_2$ has been simply set to 1 in experiments. Given $\Delta \xi$, the current pose estimate is computed as  $T^{t}=exp(\Delta \xi) \cdot T^{t-1}$, as described in Sec. \ref{sec:se3}. The details of the network architecture will be described in the released code.

\label{sec:syn_data_gen}
\subsection{Synthetic Data Generation via PPDR}

Domain randomization provides variability at training time, such that at test time, the model is able to generalize to real-world data \cite{tobin2017domain}. Prior work implements the idea of domain randomization by directly sampling object poses from a predetermined distribution \cite{tobin2017domain,xiang2017posecnn,tremblay2018deep}, which can lead to unrealistic penetration or pile configurations, adversely introducing undesired bias to depth data distribution during learning. This work therefore leverages the complementary attributes of domain randomization and physically-consistent simulation for the synthetic data generation process, namely PPDR (Physically Plausible Domain Randomization). Specifically, object poses are initialized randomly where collision between objects or distractors could occur, which is then followed by a number of physics simulation steps so that objects are separated or fall onto the table without collision. Meanwhile, other complex or intractable physical properties such as lighting, number of objects, distractor textures are randomized. For each scene, a pair of data $I^{t-1}_{syn,train}$ and $I^{t}_{syn,train}$ are generated. It is later utilized as the input to the network for training. Both images are cropped and resized into a fixed resolution $176 \times 176$ before feeding into the network - similar to prior work \cite{li2018deepim}.

The next step is to bridge the domain gap of depth data via {\it bidirectional alignment}, operated over the synthetic depth data during training time and real depth data during test time. Specifically, during training time, random Gaussian noise and random pixel corruption augmentation steps are applied to the synthetic depth data $D^{t}_{syn,train}$ at branch B to resemble a real noisy depth image. In contrast, during test time, bilateral filtering is carried out on the real depth image so as to smooth sensor noise and fill holes to be aligned with the synthetic domain.

\section{Experiments}
This  section  evaluates  the  proposed  approach  and  compares  against  state-of-the-art 6D pose tracking methods as well as single-image pose estimation methods on a public benchmark. It also introduces a new benchmark developed as part of this work, which corresponds to robot manipulation scenarios. Experiments  are  conducted  on  a  standard  desktop with  Intel  Xeon(R) E5-1660 v3@3.00GHz  processor with a NVIDIA Tesla K40c GPU. Both quantitative and qualitative results demonstrate $se(3)$-TrackNet achieves the state-of-art results in terms of both accuracy and speed, while using only synthetic training data. 


\begin{figure*}[h]
\begin{minipage}{\textwidth}
  \begin{minipage}{0.7\textwidth}
    \centering
    \resizebox{\textwidth}{!}{
\begin{tabular}{|c|cc|cc|cc|cc|cc|cc|aa||cc|aa|aa|}
\hline
                          & \multicolumn{2}{c|}{DOPE \cite{tremblay2018deep}}      & \multicolumn{2}{c|}{DenseFusion \cite{wang2019densefusion}} & \multicolumn{2}{c|}{PoseCNN+DeepIM \cite{li2018deepim}} & \multicolumn{2}{c|}{DeepIM tracking \cite{li2018deepim}} & \multicolumn{2}{c|}{RGF \cite{issac2016depth}}         & \multicolumn{2}{c|}{Wüthrich's \cite{Wthrich2013ProbabilisticOT}}  & \multicolumn{2}{a||}{$se(3)$-TrackNet}        & \multicolumn{2}{c|}{PoseRBPF \cite{deng2019poserbpf}} & \multicolumn{2}{a|}{$se(3)$-TrackNet}     & \multicolumn{2}{a|}{$se(3)$-TrackNet}     \\ \hline
Modality                  & \multicolumn{2}{c|}{RGB}       & \multicolumn{2}{c|}{RGBD}        & \multicolumn{2}{c|}{RGBD}               & \multicolumn{2}{c|}{RGB}            & \multicolumn{2}{c|}{Depth}       & \multicolumn{2}{c|}{Depth}       & \multicolumn{2}{a||}{RGBD}        & \multicolumn{2}{c|}{RGBD}     & \multicolumn{2}{a|}{RGBD}     & \multicolumn{2}{a|}{RGBD}     \\
Type                      & \multicolumn{2}{c|}{detection} & \multicolumn{2}{c|}{detection}   & \multicolumn{2}{c|}{detection}          & \multicolumn{2}{c|}{tracking}        & \multicolumn{2}{c|}{tracking}    & \multicolumn{2}{c|}{tracking}    & \multicolumn{2}{a||}{tracking}    & \multicolumn{2}{c|}{tracking} & \multicolumn{2}{a|}{tracking} & \multicolumn{2}{a|}{tracking} \\
Initial pose from         & \multicolumn{2}{c|}{-}         & \multicolumn{2}{c|}{-}           & \multicolumn{2}{c|}{-}                  & \multicolumn{2}{c|}{groundtruth}     & \multicolumn{2}{c|}{groundtruth} & \multicolumn{2}{c|}{groundtruth} & \multicolumn{2}{a||}{groundtruth} & \multicolumn{2}{c|}{PoseCNN}  & \multicolumn{2}{a|}{PoseCNN}  & \multicolumn{2}{a|}{PoseCNN}  \\
Re-initialization (Total) & \multicolumn{2}{c|}{No}        & \multicolumn{2}{c|}{No}          & \multicolumn{2}{c|}{No}                 & \multicolumn{2}{c|}{Yes (290)}       & \multicolumn{2}{c|}{No}          & \multicolumn{2}{c|}{No}          & \multicolumn{2}{a||}{No}          & \multicolumn{2}{c|}{Yes (2)}  & \multicolumn{2}{a|}{No}       & \multicolumn{2}{a|}{Yes (2)}  \\ \hline
Train data                & \multicolumn{2}{c|}{Syn}       & \multicolumn{2}{c|}{Real+Syn}    & \multicolumn{2}{c|}{Real+Syn}           & \multicolumn{2}{c|}{Real+Syn}        & \multicolumn{2}{c|}{-}           & \multicolumn{2}{c|}{-}           & \multicolumn{2}{a||}{Syn}         & \multicolumn{2}{c|}{Syn}      & \multicolumn{2}{a|}{Syn}      & \multicolumn{2}{a|}{Syn}      \\ \hline
Objects                   & ADD            & ADD-S         & ADD            & ADD-S           & ADD                & ADD-S              & ADD               & ADD-S            & ADD              & ADD-S         & ADD             & ADD-S          & ADD             & ADD-S          & ADD           & ADD-S         & ADD           & ADD-S         & ADD           & ADD-S         \\ \hline
002\_master\_chef\_can    &                & -             & -              & 96.40           & 78.00              & 96.30              & 89.00             & 93.80            & 46.23            & 90.17         & 55.60           & 90.68          & 93.86           & 96.29          & 90.50         & 95.10         & 93.84         & 95.92         & 93.84         & 95.92         \\
003\_cracker\_box         & 55.90          & 69.80         & -              & 95.50           & 91.40              & 95.30              & 88.50             & 93.00            & 56.95            & 72.26         & 96.38           & 97.19          & 96.52           & 97.20          & 88.20         & 93.00         & 96.42         & 97.12         & 96.42         & 97.12         \\
004\_sugar\_box           & 75.70          & 87.10         & -              & 97.5            & 97.60              & 98.20              & 94.30             & 96.30            & 50.38            & 72.65         & 97.14           & 97.94          & 97.58           & 98.14          & 92.90         & 95.50         & 97.56         & 98.13         & 97.56         & 98.13         \\
005\_tomato\_soup\_can    & 76.10          & 85.10         & -              & 94.60           & 90.30              & 94.80              & 89.10             & 93.20            & 72.44            & 91.60         & 64.74           & 89.55          & 94.96           & 97.17          & 90.00         & 93.80         & 94.81         & 97.10         & 94.81         & 97.10         \\
006\_mustard\_bottle      & 81.90          & 90.90         & -              & 97.20           & 97.10              & 98.00              & 92.00             & 95.10            & 87.71            & 98.19         & 97.12           & 97.95          & 95.76           & 97.37          & 91.90         & 96.30         & 95.73         & 97.36         & 95.73         & 97.36         \\
007\_tuna\_fish\_can      & -              & -             & -              & 96.60           & 92.20              & 98.00              & 92.00             & 96.40            & 28.67            & 52.93         & 69.14           & 93.32          & 86.46           & 91.09          & 91.10         & 95.30         & 86.46         & 91.08         & 86.46         & 91.08         \\
008\_pudding\_box         & -              & -             & -              & 96.50           & 83.50              & 90.60              & 80.10             & 88.30            & 12.69            & 17.98         & 96.85           & 97.89          & 97.93           & 98.39          & 85.80         & 92.00         & 97.90         & 98.37         & 97.90         & 98.37         \\
009\_gelatin\_box         & -              & -             & -              & 98.10           & 98.00              & 98.50              & 92.00             & 94.40            & 49.10            & 70.72         & 97.46           & 98.37          & 97.81           & 98.42          & 96.30         & 97.50         & 97.74         & 98.46         & 97.74         & 98.46         \\
010\_potted\_meat\_can    & 39.40          & 52.40         & -              & 91.30           & 82.20              & 90.30              & 78.00             & 88.90            & 44.09            & 45.57         & 83.71           & 86.69          & 77.81           & 84.16          & 68.70         & 77.90         & 36.45         & 60.28         & 74.51         & 82.38         \\
011\_banana               & -              & -             & -              & 96.60           & 94.90              & 97.60              & 81.00             & 90.50            & 93.33            & 97.74         & 86.27           & 96.07          & 94.90           & 97.18          & 74.20         & 86.90         & 40.04         & 78.81         & 84.62         & 95.15         \\
019\_pitcher\_base        & -              & -             & -              & 97.10           & 97.40              & 97.90              & 90.40             & 94.70            & 97.93            & 98.18         & 97.30           & 97.74          & 96.75           & 97.45          & 86.80         & 94.20         & 96.71         & 97.43         & 96.71         & 97.43         \\
021\_bleach\_cleanser     & -              & -             & -              & 95.80           & 91.60              & 96.90              & 81.70             & 90.50            & 95.87            & 97.28         & 95.23           & 97.16          & 95.94           & 97.25          & 86.00         & 93.00         & 95.89         & 97.23         & 95.89         & 97.23         \\
024\_bowl                 & -              & -             & -              & 88.20           & 8.10               & 87.00              & 38.80             & 90.60            & 24.25            & 82.40         & 30.37           & 97.15          & 80.91           & 94.46          & 25.50         & 94.20         & 39.12         & 95.56         & 39.12         & 95.56         \\
025\_mug                  & -              & -             & -              & 97.10           & 94.20              & 97.60              & 83.20             & 92.00            & 59.99            & 71.18         & 83.15           & 93.35          & 91.53           & 96.88          & 90.90         & 97.10         & 91.56         & 96.88         & 91.56         & 96.88         \\
035\_power\_drill         & -              & -             & -              & 96.00           & 97.20              & 97.90              & 85.40             & 92.30            & 97.94            & 98.35         & 97.09           & 97.82          & 96.42           & 97.40          & 93.90         & 96.10         & 96.38         & 97.38         & 96.38         & 97.38         \\
036\_wood\_block          & -              & -             & -              & 89.70           & 81.10              & 91.50              & 44.30             & 75.40            & 45.68            & 62.51         & 95.48           & 96.87          & 95.16           & 96.70          & 20.10         & 89.10         & 33.91         & 95.92         & 33.91         & 95.92         \\
037\_scissors             & -              & -             & -              & 95.20           & 92.70              & 96.00              & 70.30             & 84.50            & 20.94            & 38.60         & 4.17            & 16.20          & 95.68           & 97.55          & 76.10         & 85.60         & 95.67         & 97.54         & 95.67         & 97.54         \\
040\_large\_marker        & -              & -             & -              & 97.50           & 88.90              & 98.20              & 80.40             & 91.20            & 12.17            & 18.90         & 35.58           & 53.02          & 92.15           & 95.99          & 92.00         & 97.10         & 89.01         & 94.23         & 89.01         & 94.23         \\
051\_large\_clamp         & -              & -             & -              & 72.90           & 54.20              & 77.90              & 73.90             & 84.10            & 62.84            & 80.12         & 61.25           & 72.35          & 94.71           & 96.93          & 48.50         & 94.80         & 71.60         & 96.88         & 71.60         & 96.88         \\
052\_extra\_large\_clamp  & -              & -             & -              & 69.80           & 36.50              & 77.80              & 49.30             & 90.30            & 67.48            & 69.65         & 93.73           & 96.58          & 91.74           & 95.76          & 40.30         & 90.10         & 64.58         & 95.80         & 64.58         & 95.80         \\
061\_foam\_brick          & -              & -             & -              & 92.50           & 48.20              & 97.60              & 91.60             & 95.50            & 69.99            & 86.55         & 96.76           & 98.11          & 93.65           & 96.71          & 81.10         & 95.70         & 40.66         & 94.67         & 40.66         & 94.67         \\ \hline
ALL                       & -              & -             & -              & 93.10           & 80.70              & 94.00              & 79.30             & 91.00            & 59.18            & 74.29         & 78.01           & 90.21          & 93.05           & 95.71          & 80.80         & 93.30         & 84.46         & 93.87         & 87.81         & 95.52         \\ \hline
Speed (fps)               & \multicolumn{2}{c|}{4.31}      & \multicolumn{2}{c|}{16.67}       & \multicolumn{2}{c|}{0.09}               & \multicolumn{2}{c|}{12.00}           & \multicolumn{2}{c|}{11.76}       & \multicolumn{2}{c|}{12.93}       & \multicolumn{2}{a||}{90.90}       & \multicolumn{2}{c|}{5.00}     & \multicolumn{2}{a|}{90.90}    & \multicolumn{2}{a|}{90.90}    \\ \hline
\end{tabular}%
}
    \end{minipage}
  \begin{minipage}{0.3\textwidth}
    \centering
    \includegraphics[width=\textwidth,height=0.45\textwidth]{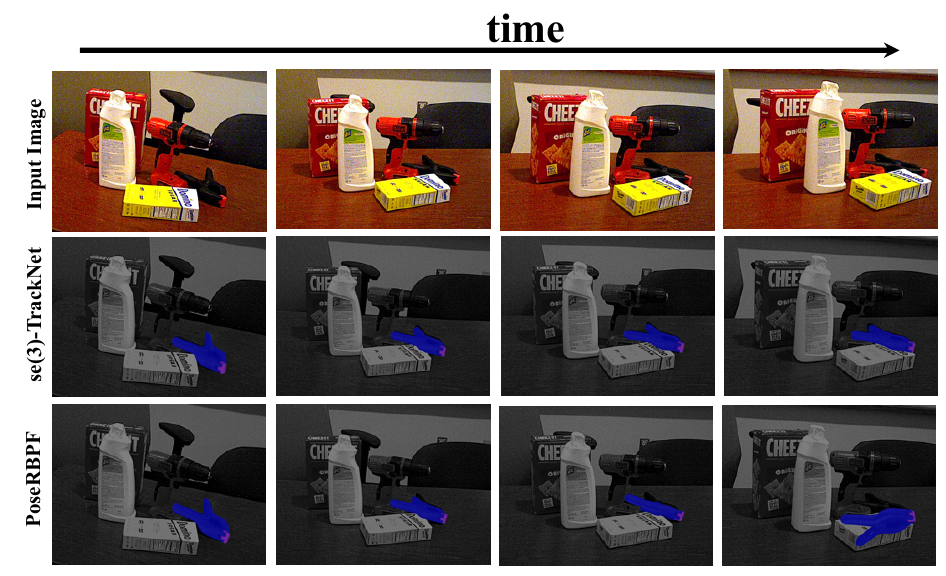}
    \includegraphics[width=\textwidth,height=0.45\textwidth]{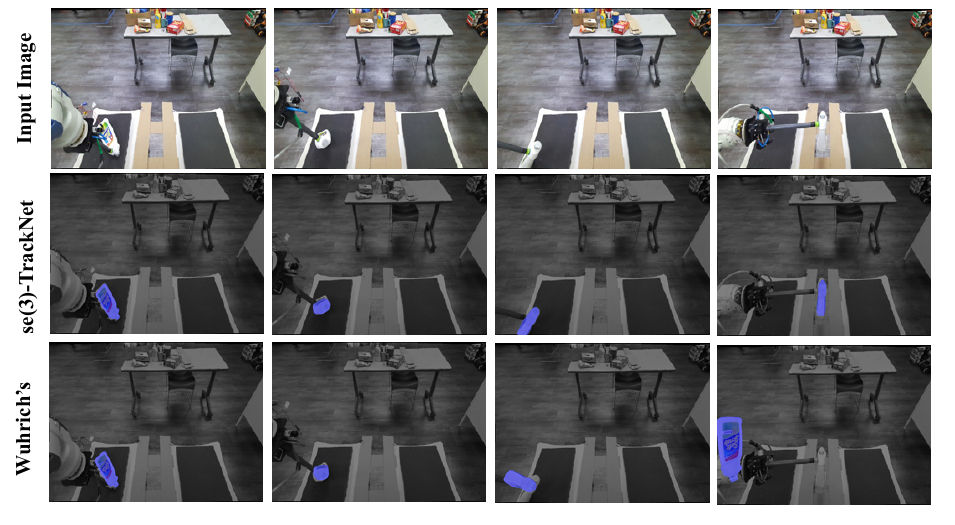}
  \end{minipage}  
\end{minipage}
\vspace{-0.1in}
\caption{ \scriptsize \textbf{Left:} Comparing the performance of $se(3)$-TrackNet (Gray) with state-of-the-art techniques on the {\it YCB Video}. The approach significantly outperforms the competing approaches over the ADD metric, which considers semantic information during pose evaluation. It also achieves the highest success rate over the ADD-S metric both in cases of initialization with the ground-truth pose and when initialized with the output of PoseCNN \cite{xiang2017posecnn} (rightmost two columns).  \textbf{Top-right:} Qualitative results for tracking the  "large-clamp" object in the YCB-Video dataset. \textbf{Bottom-right:} Tracking results for "bleach-cleanser" being manipulated by a vacuum gripper in the {\it YCBInEOAT} dataset.} \label{fig:ycb_video}
\vspace{-0.2in}
\end{figure*}

\subsection{Datasets}
\noindent \textbf{YCB-Video Dataset} This dataset \cite{xiang2017posecnn} captures 92 RGB-D video sequences over 21 YCB Objects \cite{calli2015ycb} arranged on table-tops. Objects' groundtruth 6D poses are annotated in every frame. The evaluation closely follows the protocols adopted in comparison methods \cite{xiang2017posecnn, deng2019poserbpf, li2018deepim, wang2019densefusion} and reports the AUC (Area Under Curve) results on the keyframes in 12 video test sequences evaluated by the metrics of $ADD = \frac{1}{m}\sum_{x \in M} ||Rx+T-(\hat{R}x+\hat{T})||$ which performs exact model matching, and $ADD-S = \frac{1}{m}\sum_{x_1 \in M} \min_{x_2 \in M}||Rx_1+T-(\hat{R}x_2+\hat{T})||$ \cite{xiang2017posecnn}. 

\noindent \textbf{YCBInEOAT Dataset} In order to evaluate the 6D tracking performance in the setup of dynamic moving object, in this work, a novel dataset, referred to as "YCBInEOAT Dataset", is developed in the context of robotic manipulation, where various robot end-effectors are included: a vacuum gripper, a Robotiq 2F-85 gripper, and a Yale T42 Hand \cite{odhner2013open}. The manipulation sequences consider 5 YCB objects outlined in Table \ref{tab:mydata_res}. Each video sequence is collected from a real manipulation performed with a dual-arm {\it Yaskawa Motoman SDA10f}. In general, there are 3 types of manipulation tasks performed: (1) single arm pick-and-place, (2) within-hand manipulation, and (3) pick to hand-off between arms to placement. RGB-D images are captured by an {\it Azure Kinect} sensor mounted statically on the robot with a frequency of 20 to 30 Hz. Similar to the YCB-Video, ADD and ADD-S metrics are adopted for evaluation. Ground-truth 6D object poses in camera's frame have been accurately annotated \emph{manually} for each frame of the video.



\subsection{Results on YCB-Video}

Fig. \ref{fig:ycb_video} present the evaluation over the {\it YCB-Video} dataset. Although this dataset contains real world annotated training data, $se(3)$-TrackNet does not use any of them but is trained solely on synthetic data generated by aforementioned pipeline. It is compared with other state-of-art 6D object pose detection approaches   \cite{xiang2017posecnn, tremblay2018deep, li2018deepim, wang2019densefusion} and 6D pose tracking approaches \cite{deng2019poserbpf, li2018deepim, issac2016depth, Wthrich2013ProbabilisticOT}, where publicly available source code\footnote{https://github.com/bayesian-object-tracking/dbot} is used to evaluate \cite{issac2016depth,Wthrich2013ProbabilisticOT}, while other results are adopted from the respective publications. All the compared tracking methods except PoseRBPF are using ground-truth pose for initialization. PoseRBPF \cite{deng2019poserbpf} is the only one that is initialized using predicted poses from PoseCNN \cite{xiang2017posecnn}. For fairness, two additional experiments using the same initial pose as PoseRBPF\footnote{We thank the authors for providing the initial poses they used in their original paper \cite{deng2019poserbpf}.} are performed and presented in the rightmost two columns of Table I, one is without any re-initialization, and the other allows re-initialization by PoseCNN twice (same as in PoseRBPF) after heavy occlusions. The prior work \cite{li2018deepim} was originally proposed to refine the pose output from any 6D pose estimation detection method, but also extends to RGB-based tracking. It has to be re-initialized by PoseCNN when the last 10 frames have an average rotation greater than 10 degrees or an average translation greater than 1 cm, which happens every 340 frames on average as reported \cite{li2018deepim}.


\subsection{Results on YCBInEOAT-Dataset}
\begin{table}[]
\centering
\resizebox{0.48\textwidth}{!}{%
\begin{tabular}{ccccccc}
\hline 
Objects & \multicolumn{2}{c}{RGF \cite{issac2016depth}} & \multicolumn{2}{c}{Wüthrich's \cite{Wthrich2013ProbabilisticOT}} & \multicolumn{2}{c}{$se(3)$-TrackNet} \\
& ADD       & ADD-S       & ADD         & ADD-S         & ADD         & ADD-S      \\
\hline                    
003\_cracker\_box     & 34.78          &  55.44           & 79.00            &   88.13            & 90.76       & 94.06      \\
021\_bleach\_cleanser &  29.40         &  45.03           &  61.47           &  68.96             & 89.58       & 94.44      \\
004\_sugar\_box       &  15.82         &  16.87           &  86.78           &   92.75            & 92.43       & 94.80      \\
005\_tomato\_soup\_can &  15.13         &  26.44           &  63.71           &   93.17            & 93.40       & 96.95      \\
006\_mustard\_bottle  &   56.49        &  60.17           &  91.31           &  95.31             & 97.00       & 97.92      \\
\hline 
ALL                 &  29.98         & 39.90            &  78.28           &   89.18            & 92.66       & 95.53\\
\hline
\end{tabular}%
}
\caption{\scriptsize Results evaluated on YCBInEOAT-dataset by AUC (Area Under Curve) for ADD and ADD-S.}
\label{tab:mydata_res}
\end{table}

Table \ref{tab:mydata_res} shows the quantitative results evaluated by the area under the curve for ADD and ADD-S on the developed {\it YCBInEOAT} dataset. On this benchmark, the tracking approaches with publicly available source code could be directly evaluated \cite{Wthrich2013ProbabilisticOT,issac2016depth}. Pose is initialized with ground-truth in the first frame and no re-initialization is allowed. Example qualitative results are demonstrated by Fig. \ref{fig:ycb_video}. Abrupt motions, extreme rotations and slippage within the end-effector expose challenges for 6D object pose tracking. Nevertheless, $se(3)$-TrackNet is sufficiently robust to perform long-term reliable pose tracking.

\label{sec:ablation}
\subsection{Ablation Study}

\begin{wrapfigure}{r}{0.5\linewidth}
\vspace{-0.1in}

\resizebox{0.245\textwidth}{!}{%
\hspace{-0.3in}
\begin{tabular}{c|c|c}
\hline
Criteria & ADD   & ADD-S \\ \hline
Proposed             & 94.71 & 96.93 \\
No physics       & 91.88 & 95.76 \\
No depth         & 75.65 & 87.22 \\
Shared encoder   & 0.28  & 0.28  \\
Quaternion       & 93.58 & 96.39 \\
Shape-Match Loss & 1.93  & 5.48  \\ \hline
\end{tabular}%
}
\label{tab:ablation}
\vspace{-0.17in}
\end{wrapfigure}
An ablation study investigates the importance of different modules of the proposed approach. It is performed for the {\it large clamp} object from the {\it YCB-Video dataset} and is presented in the accompanying table. The initial pose is given by ground-truth and no re-initialization is allowed. \textbf{No physics} implies that physics simulation is removed during synthetic data generation. For \textbf{No depth}, the depth modality is removed in both training and inference stage. \textbf{Shared encoder} means $I^t$ and $I^{t-1}$ are passed to the same feature encoder. \textbf{Quaternion} replaces the proposed rotation representation by quaternion. \textbf{Shape-Match Loss} \cite{xiang2017posecnn} does not require the specification of symmetries. However, it loses track of the object early in the video.

{\small
\bibliographystyle{ieee_fullname}
\bibliography{egbib}
}

\end{document}